\documentclass{esannV2}
\usepackage[dvips]{graphicx}
\usepackage{overpic}
\usepackage[latin1]{inputenc}
\usepackage{amssymb,amsmath,array}
\usepackage[T1]{fontenc}
\usepackage{svg}
\usepackage{xcolor}
\usepackage{multirow}

\usepackage[backend=biber,sortcites=true,sorting=none,maxnames=100,minnames=1,style=ieee,citestyle=numeric-comp]{biblatex}
\usepackage[font=small,labelfont=bf]{caption}

\usepackage{fancyhdr}
\fancypagestyle{firstpage}{%
  \fancyhf{}
  \fancyfoot[L]{\small\itshape Preprint. To appear in the Proceedings of the European Symposium on Artificial Neural Networks (ESANN), 2026.}

}

\begin{document}

\title{TSFM in-context learning for time-series classification of bearing health status}

\author{Michel Tokic$^{1,2}$, Slobodan Djukanovi\'{c}$^{1,3}$, Anja von Beuningen$^1$, Cheng Feng$^1$ 
%
%
\vspace{.3cm}\\
%
1- Siemens AG, Data \& Artificial Intelligence, Otto-Hahn-Ring 6, \\81739 Munich, Germany\vspace{.1cm}\\
2- Faculty of Mathematics, Informatics and Statistics, \\Ludwig-Maximilians-University Munich, Munich, Germany\vspace{.1cm}\\
3- Faculty of Electrical Engineering, University of Montenegro, \\Podgorica, Montenegro
}
\maketitle
\thispagestyle{firstpage} 

\begin{abstract}
We introduce a classification method based on in-context learning using time-series foundation models (TSFMs). We demonstrate how data not included in the TSFM training can be classified without fine-tuning the foundation model or training a traditional classification model. Examples are represented as targets (class labels) and covariates (data matrices) within the TSFM prompt, enabling the classification of unknown covariate data patterns alongside the forecast horizon through in-context learning.
We apply this method to vibration data to assess the health state of a bearing within a servo-press motor. The method transforms frequency-domain reference signals into pseudo time-series patterns, generates aligned covariate and target signals, and uses the TSFM to predict class-membership probabilities for predefined labels.
Leveraging the scalability of pre-trained models, the proposed method demonstrates effectiveness across varying operational conditions. This represents significant progress beyond traditional, custom AI solutions towards broader AI-driven  maintenance systems that could potentially be provided as Model- or Software-as-a-Service applications.
\end{abstract}

\section{Introduction}
In recent years, industries such as manufacturing, energy, transportation, and healthcare have increasingly recognized the importance of anomaly detection to prevent failures of critical machines and systems. The traditional maintenance approach involves reactive strategies that can lead to costly downtime, production losses, and safety risks. As a result, many companies are shifting towards predictive and preventive maintenance strategies, aiming to optimize operational efficiency by preemptively addressing signs of equipment deterioration.

Despite advancements in predictive maintenance, current solutions are largely tailored for specific applications and assets, relying heavily on domain-specific knowledge and handcrafted rules \cite{qiu2023deep}. Although effective in controlled contexts, these systems often lack flexibility and require substantial effort to adapt to new environments or equipment.

The rise of artificial intelligence (AI) has introduced AI-driven maintenance strategies that rely on data-driven approaches. However, existing implementations primarily depend on traditional AI solutions, trained on asset-specific datasets, and typically struggle to scale across different machines, industries, or failure patterns \cite{qiu2023deep}. This limitation is due to their inability to generalize beyond predefined conditions, forcing companies to rely on custom-built solutions that hinder the full potential of AI-driven maintenance.

This work addresses these challenges by introducing a novel approach for the classification of data, specifically vibration data, using a time-series foundation model (TSFM).
Unlike traditional AI models, TSFMs offer the generalization capabilities required for scalable and flexible classification solutions across different assets and operating conditions. Notably, they enable reliable data classification without extensive fine-tuning of large reference datasets or the need to train a traditional AI classification model \cite{auer2025pretrainedforecastingmodelsstrong}. This represents a significant step towards adaptive and generalizable AI models \cite{liang2024foundation} and could potentially be offered to customers as Model- or Software-as-a-Service applications.

\section{Method}
This chapter details the proposed methodology for classifying the health state of a motor bearing using spectral data.

\subsection{Time Series Foundation Models}
TSFMs have emerged as a transformative paradigm for time-series forecasting, delivering strong zero-shot and few-shot performance across a broad spectrum of datasets and application domains \cite{woo2024unified,das2024decoder,ansari2025chronos}. Recent studies demonstrate that TSFMs can match or even surpass specialized statistical models and earlier task-specific models on unseen benchmarks. Our method is centered on the \textit{General Time Transformer} (GTT) architecture \cite{feng2024general} that captures complex temporal patterns and multivariate dependencies through alternating temporal and channel attentions in Transformer encoder blocks. By adding a learnable sink token to the end of channels for target variates similar to \cite{woo2024unified}, and replacing the original point forecast head with a probabilistic forecast head that uses a four-component Gaussian Mixture Model to calibrate the output distribution, we adapt the GTT architecture for multivariate probabilistic forecast with covariates that are available in the forecast horizon \cite{tokic2020handling}.
Pretrained on a large-scale cross-domain dataset, GTT is a general-purpose zero-shot forecaster that takes a look-back window of $L$ time points for one or more target variables (optionally with covariates) and predicts their next $H$ values. Let $\mathbf{x}_{1:L}$ and $\mathbf{c}_{1:L+H}$ be the target variates and corresponding covariates. Targets are available only in the context window, whereas the covariates can be available also in the forecast horizon. The model is a function that predicts the distribution of $\mathbf{x}_{L+1:L+H}$, such that $ P(\mathbf{x}_{L+1:L+H}) = f( \mathbf{x}_{1:L}, \mathbf{c}_{1:L+H} )$.

\begin{figure}[t!]
     \centering
        \begin{minipage}{0.22\textwidth}
            \vspace{-6pt}
            \centering
            \includegraphics[width=\textwidth]{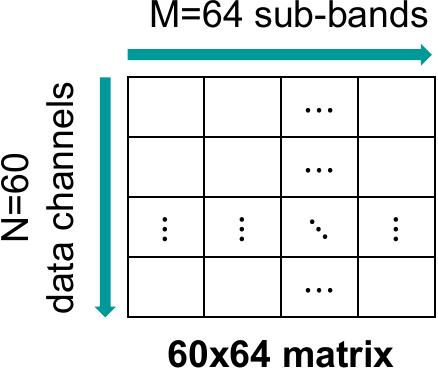}    
        \end{minipage}
    \hfill
    \begin{minipage}[r]{0.75\textwidth}
            \vspace{0pt}
            \centering
            \includegraphics[width=\textwidth]{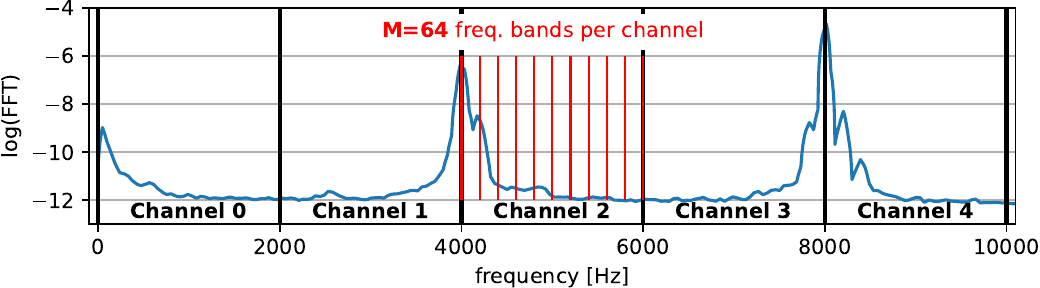}
    \end{minipage}
     
     \caption{Preprocessing: Signal spectrum (magnitude of FFT) is transformed into a $60\times64$ matrix ($N=60$ data channels and $M=64$ frequency sub-bands per channel). \textit{Note:} On the right graph we only depict five out of 60 possible channels. }
     \label{fig:data_preprocessing}
\end{figure}

\begin{figure}[b!]
     \centering
     \includegraphics[width=\textwidth]{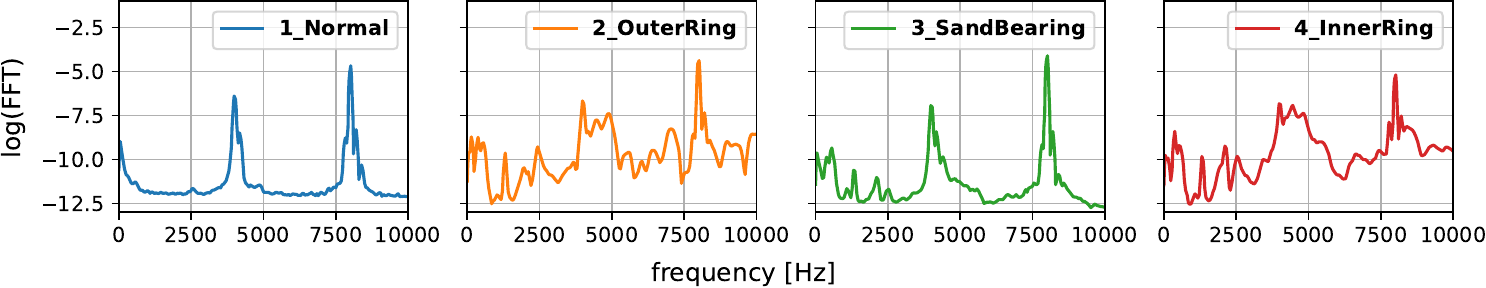} 
     \caption{Example FFTs from the servo-press dataset. Note similarity between 
     \textit{1\_Normal} and \textit{3\_SandBearing} and between \textit{2\_OuterRing} and \textit{4\_InnerRing}.}
     \label{fig:examples}
\end{figure}

\subsection{Data preprocessing}
Raw vibration signals sampled at 48 kHz from a servo-press motor are processed to compute the fast Fourier transform (FFT) \cite{steven1993fundamentals}, with recordings' length of $\approx$ 60 seconds. 
As illustrated in Fig.~\ref{fig:data_preprocessing}, the preprocessing divides the whole spectrum into \textit{$N=60$ data channels} (covariates). For each channel, the FFT is segmented into \textit{$M=64$ frequency sub-bands}, with the mean value from each sub-band forming a feature. Overall, one raw vibration signal results in a covariate structure of size $N \times M$, used as input to the TSFM. No additional covariates have been added.

\begin{figure}[t!]
     \centering
     \begin{overpic}[width=\textwidth]{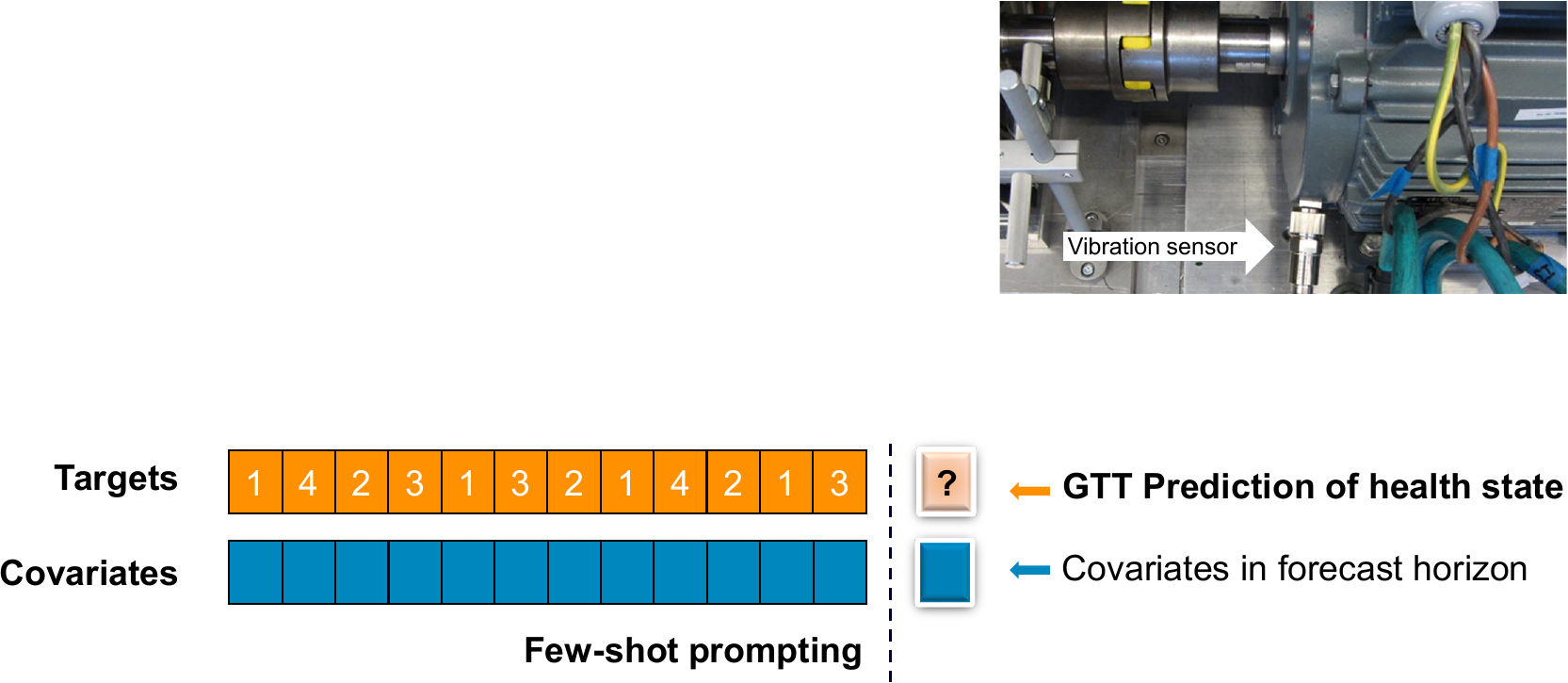}
     \put(0, 17.7){\includegraphics[width=0.6035\linewidth]{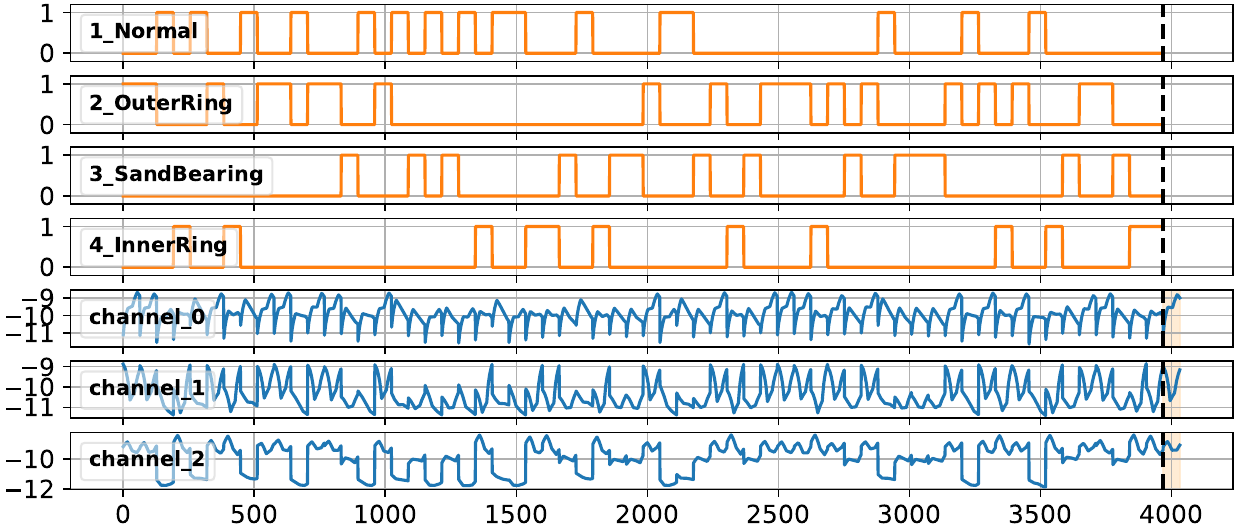}}
     \end{overpic}
     \caption{A sequence of random samples from the servo-press dataset with one-hot encoded targets (orange) and corresponding covariates (blue) forms the few-shot prompting context. The model outputs a prediction of the targets w.r.t. covariates in the forecast horizon (blue block to the right of the dashed line). \textit{Note:} Only the first three out of 60 covariate channels are depicted in the upper context.}
     \label{fig:few_shot_prompting}
\end{figure}

The system defines target variables for classification, corresponding to four health states (classes): 1. \textit{Normal operation}, 2. \textit{Outer ring fault}, 3. \textit{Sand in bearing}\footnote{Sand has been artificially added in the bearing.}, and 4. \textit{Inner ring fault}. Targets are one-hot encoded (1 if the class is present, 0 otherwise) over a forecast horizon of $M=64$ time steps, enabling sensitivity to subtle variations and emphasizing fault-indicative frequency components (see Figs. \ref{fig:data_preprocessing}-\ref{fig:few_shot_prompting}).

\subsection{Few-shot prompting for in-context learning classification}
We apply a few-shot prompting to adapt the pre-trained GTT for asset health classification tasks, as shown in Fig.~\ref{fig:few_shot_prompting}.

A small sequence of recorded target health states and their corresponding covariates forms the "context" or "prompt". This allows the model to learn about patterns associated with specific health states from a limited number of examples, which is known as \textit{in-context learning} or \textit{few-shot prompting}.

For a new covariate input, the model predicts the corresponding class as depicted in Fig.~\ref{fig:result}. Within 64 time steps, the model predicts the class of the tested covariate data. As a classification method, we apply a winner-takes-all rule at the last forecast step $\mathbf{x}_{L+H}$; alternatively, a discrete probability distribution could be obtained by applying a Softmax function to the predicted intensities.

\section{Experiments}

We conducted experiments with a GTT model with 750 million parameters pretrained on a large-scale cross domain time series corpus with 124 billion data points. The model's context accommodates 64 channels/variates, with a maximum of 4480 time steps, which is sufficient for up to $4480/64=70$ servo-press data samples (vibration signals). Each sample occupies $M=64$ steps and $N=60$ covariate channels, as shown in Fig.~\ref{fig:data_preprocessing} (left).

The dataset comprises 280 nearly class-balanced samples depicted in Fig.~\ref{fig:examples}. 
Including four target variables, we have a total of 64 variables (targets + covariates) per class sample, which corresponds to the current limit of GTT.

\begin{figure}[t!]
     \centering
     \includegraphics[width=\textwidth]{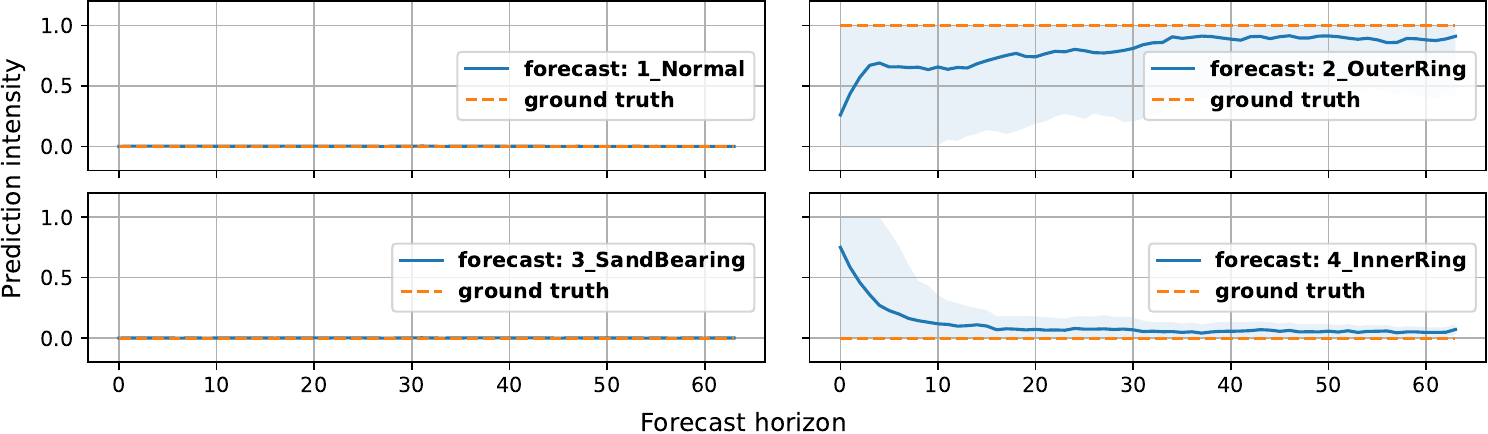} 
     \caption{Correct classification of class \textit{2\_OuterRing} (upper right), based on corresponding covariate matrix available in the GTT forecast horizon. Note, the prediction intensity stays/develops towards 0 for all other classes.}
     \label{fig:result}
\end{figure}

A typical classification is depicted in Fig.~\ref{fig:result}, where the task was to predict class \textit{2\_OuterRing}. In the corresponding prompt, the last example was drawn from class \textit{4\_InnerRing}, explaining why the prediction for this class starts with high intensity (around 0.7) at the beginning, but develops down to almost 0 until step $t=63$. In contrast, the intensity of class \textit{2\_OuterRing} rises towards almost 1. At $t=63$, the applied winner-takes-it all rule classifies the data correctly.

\begin{figure}[b!]
     \centering
     \includegraphics[width=\textwidth]{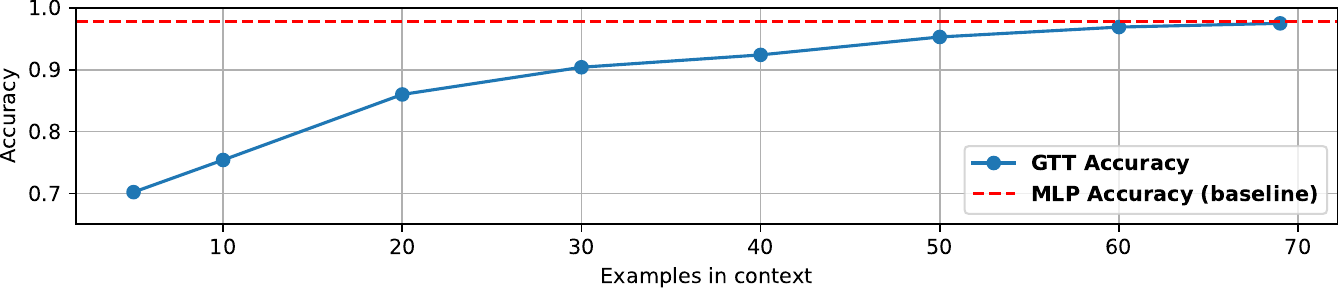} 
     \caption{Comparison of classification accuracy over varied amount of few-shot examples. }
     \label{fig:result_comparison_mlp}
\end{figure}

An extensive study has been carried out to analyze the dependency on the number of context examples. Figure~\ref{fig:result_comparison_mlp} presents the classification accuracy, obtained over 1000 runs, for randomly selected few-shot examples with varying context sizes (number of examples). Random sampling enables varying the last sample in the context, just before the forecast is generated (see Fig.~\ref{fig:result}).

A classification accuracy of 97.5\,\% is achieved when using the full context length of 4480 time-steps. All four classes were identified with high precision and recall, exhibiting only minor misclassifications. These results demonstrate that the TSFM-based approach is highly reliable and consistent in differentiating between normal and faulty bearing conditions, particularly considering that this dataset was excluded from the model's training data.

As a baseline, a multilayer perceptron (MLP) classifier was used, consisting of two fully connected hidden layers with 256 and 32 neurons, each followed by ReLU activation and 0.4 dropout, and a final softmax four-class output layer. The input spectra were preprocessed using the same aggregation procedure as in the proposed GTT method, but provided as a single one-dimensional feature vector rather than separated into covariates. Performance was evaluated using repeated stratified random splits (70/15/15 train-validation-test) with different random seeds in each run. Averaged over 20 runs, the MLP achieved an accuracy of 97.9\,\% (dashed red line in Fig.~\ref{fig:result_comparison_mlp}).

\section{Results and Discussion}
The presented methodology offers significant advantages over traditional AI approaches by combining pre-trained TSFMs with few-shot prompting. It enables rapid deployment, eliminates training time, minimizes expert dependency, and ensures scalability across diverse assets and failure modes. 
Our work highlights the broader potential of TSFMs to scale to entire new applications, such as classification as investigated in our work. The method's flexibility makes it adaptable to diverse assets and fault types, or even to entire new classification tasks.

The results demonstrate that aligning covariates and target data as pseudo time-series patterns effectively utilized the TSFM's generalization capacity. The context with 69 few-shot examples yielded 97.5\,\% accuracy, on-par performance with the MLP baseline (97.9\,\%). Both consistently distinguish between normal and faulty conditions, although the servo-press dataset was not part of the GTT training data. 
The FFT-based approach was crucial in highlighting fault-indicative frequencies, and the presentation of more random examples in the context contributed to minimizing the confusion between classes.

While this method is promising, a current limitation is the application to only a few target classes, such as four classes as in our study. 
Since the context length in the examined GTT model is currently limited to 4480 time steps, a higher number of classes would result in less represented examples per class, with potentially negative impact on the classification accuracy.


\printbibliography

\end{document}